\documentclass{article}

\usepackage{microtype}
\usepackage{graphicx}
\usepackage{subcaption}
\usepackage{booktabs} 
\usepackage{listings} 
\usepackage[breaklinks=true]{hyperref}
\usepackage{alltt}

\usepackage{hyperref}
\newcommand{\head}[1]{\noindent \textbf{#1}}

\usepackage[accepted]{sysml2019}

\usepackage{xcolor}
\definecolor{bluekeywords}{rgb}{0.13,0.13,1}
\definecolor{greencomments}{rgb}{0,0.5,0}
\definecolor{greynumber}{rgb}{0.6,0.6,0.6}
\definecolor{redstrings}{rgb}{0.9,0,0}
\sysmltitlerunning{Wield: Systematic Reinforcement Learning With Progressive Randomization}

\begin{document}

\twocolumn[
\sysmltitle{Wield: Systematic Reinforcement Learning \\ With Progressive Randomization}



\sysmlsetsymbol{equal}{*}

\begin{sysmlauthorlist}
\sysmlauthor{Michael Schaarschmidt}{cam}
\sysmlauthor{Kai Fricke}{hhsu}
\sysmlauthor{Eiko Yoneki}{cam}
\end{sysmlauthorlist}

\sysmlaffiliation{cam}{University of Cambridge}
\sysmlaffiliation{hhsu}{Helmut-Schmidt-University Hamburg}

\sysmlcorrespondingauthor{Michael Schaarschmidt}{michael.schaarschmidt@cl.cam.ac.uk}

\sysmlkeywords{Machine Learning, SysML}

\vskip 0.3in

\begin{abstract}
Reinforcement learning frameworks have introduced abstractions to implement and execute algorithms at scale. They assume standardized simulator interfaces but are not concerned with  identifying suitable task representations. We present Wield, a first-of-its kind system to facilitate task design for practical reinforcement learning. Through software primitives, Wield enables practitioners to decouple system-interface and deployment-specific configuration from state and action design. To guide experimentation, Wield further introduces a novel task design protocol and classification scheme centred around staged randomization to incrementally evaluate model capabilities.
\end{abstract}
]



\printAffiliationsAndNotice{}  

\section{Introduction}
\label{introduction}
Following high profile successes in domains like games and robotics, interest in applications of deep reinforcement learning (RL) has seen explosive growth. In computer systems, RL has found applications across a diverse range of domains  such as scheduling \cite{Mao2019}, networking \cite{Valadarsky2017}, database management  \cite{Marcus2018, 2019arXiv190403711M}, and device placement optimization \cite{mirhoseini2017device, hierarchical2018}. 

Essential for the proliferation of RL applications have been open source implementations of popular algorithms. Algorithmic frameworks like RLlib \cite{Liang2019}, RLgraph \cite{Schaarschmidt2019} or OpenAI baselines \cite{baselines} allow practitioners to execute off-the-shelf algorithms at scale. These frameworks standardize task execution through shared interfaces such as OpenAI gym \cite{openaigym}. They do not concern themselves with identifying problem representations.

In consequence, RL applications in systems have seen limited standardization. While a multitude of  experimental successes have been reported in controlled environments, real-world data processing systems are yet to widely utilize RL. Experimental research often relies on highly customized benchmarks, hardware setups, state-action representations, and proprietary simulators. Moreover, assessing evaluations  is complicated by the use of fixed workloads and limited reporting on the impact of random seeds and workload variation. Hence, applied research is fragmented, and novel approaches are difficult to reproduce. Their viability across larger deployments or different tasks remains unclear.

RL algorithms suffer from known limitations due to large sample requirements, sensitivity to hyper-parameters \cite{Henderson2017}, random weight initialization, and small input perturbations \cite{Kansky2017}. In this paper, we argue that another root cause of limited real-world progress is a lack of shared evaluation protocols and design tools. Specifically, both reinforcement learning agents and systems workloads can exhibit several degrees of non-determinism.

For example, RL agents have been used in blackbox optimization settings where an agent is trained to optimize a single fixed workload instance (e.g. a single computation graph \cite{hierarchical2018}). Within a blackbox setting,  experiments with varied random seeds can each perform on the same fixed task (fixed blackbox), or sample a random task instance per experiment to illustrate robustness to task variation (randomized blackbox). Similar evaluation modes can be applied to generalization problems, with additional consideration for within- and out-of-distribution evaluation.

To begin addressing these difficulties, we present Wield, a first-of-its-kind system towards systematic task design and evaluation in applied RL. Wield makes two contributions:

First, Wield provides a small set of reusable software primitives which decouple system interface and RL representation from deployment-specific data and task layouts. These primitives are coordinated through standardized workflows which help researchers explore new state, action, and reward models independent of system-specifics.

Second, we introduce \textbf{progressive randomization},  a novel task evaluation protocol  and classification scheme which explicitly delineates sources of non-determinism. Progressive randomization enables practitioners to communicate evaluation assumptions, and to incrementally evaluate model capabilities.

In the remainder of the paper, we first introduce Wield's design abstractions and discuss common issues around model design in systems-RL (RL  applied to systems). We then introduce progressive randomization and use it to review recent work in systems-RL. In the evaluation, we demonstrate Wield's utility by reviewing the device placement problem, a popular task in systems-RL. We reproduce prior work to classify its capabilities through progressive randomization, and subsequently implement a novel placer with Wield. Our results illustrate the true cost of evaluating RL solutions, and further call into question common evaluation practices on fixed datasets.

\section{Wield}
\subsection{Overview}
Delineating practical progress requires systematic assessment and comparison of approaches. The aim of Wield is to provide re-usable abstractions to standardize task design for systems applications of reinforcement learning. 

\begin{figure}[t] 
\centering
\includegraphics[scale=0.5]{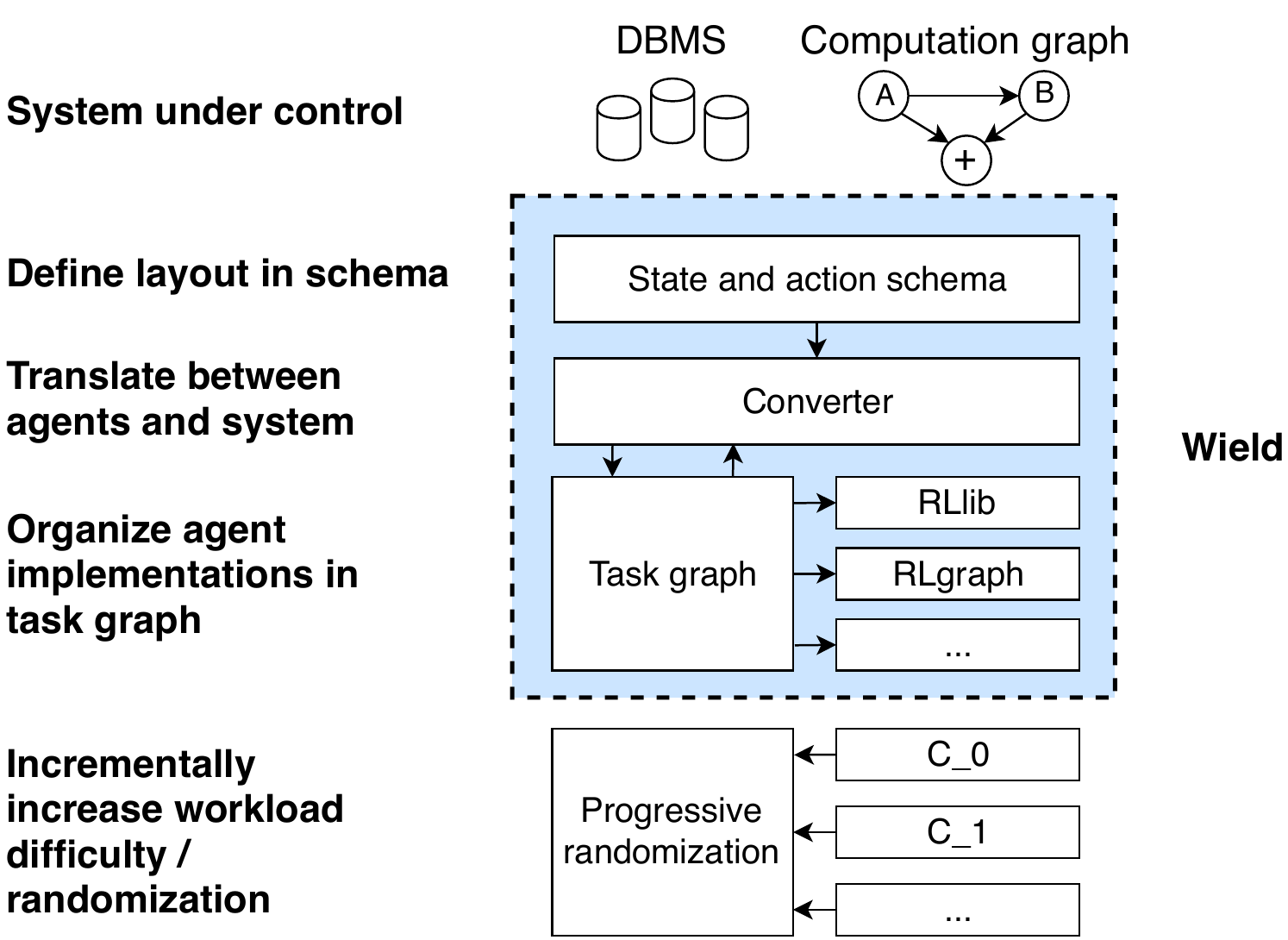}
\caption{Wield overview. To interface a system, users first implement a schema specifying data layouts. Next, a converter uses the schema to implement a mapping between agent and system view. Finally, Wield coordinates RL-agents arranged in a task graph. Progressive randomization guides incremental evaluation.}
\label{fig:wield-overview}
\vspace{-5mm}
\end{figure} 
Figure \ref{fig:wield-overview} gives  a conceptual overview of Wield. On a high level, Wield acts as an interface between a data-processing system (e.g. database, scheduler, distributed execution engine) and a reinforcement learning framework such as RLgraph or RLlib, auto-tuners, or any implementation exposing a task interface. We implemented our Wield prototype using RLgraph's agent implementations  \cite{Schaarschmidt2019}. The highest level abstraction in Wield are \textit{workflows} which coordinate execution of online (interacting with a system) or offline (e.g. log data) training, evaluation, and serialization of data and models.

Models use \textit{task graphs} to describe hierarchies of tasks wherein a single node may be  a single differentiable agent architecture, a blackbox auto-tuner, or a supervised model. Tasks use \textit{converters} to map between agent and system view of data, and \textit{schemas} to standardize programmatic layouts of input states and actions. By separating representation design and system-specifics, task architectures can be used across similar systems or problem structures which only differ in the system interface (e.g. different databases with distinct query languages). 

\subsection{Task design abstractions}
Wield's task abstractions unify workflow streams via standardized task and data layouts. Layout refers to the concrete dimensions, data types and processing steps for all inputs and outputs of an RL agent. As a running example, we use the popular problem of placing a computational graph (e.g. a TensorFlow graph) across a heterogeneous set of devices to minimize runtime of the update operation \cite{mirhoseini2017device, hierarchical2018,Addanki2019}.

\subsection{Task schemas}
\head{State design.} Task schemas are motivated by the observation that states for systems problems need to be explicitly designed. In contrast, game simulators like Atari have fixed state dimensions (e.g. 640 $\times$ 480 images) across games. All methods can rely on a fixed base representation (i.e. the original game frame) for reproducible and comparable experiments. Wield schemas encapsulate input-dependent state and action layout construction. 

Consider a state model for encoding an operation in a computational graph. The state may encode various types of semantic problem information such as operator type embeddings, tensor shapes, current device, and topology of the local graph neighborhood \cite{hierarchical2018}. While iteratively exploring different representations, schemas capture the layout of the resulting states. Moreover, different resource types and layouts, i.e. dimensions of state arrays, may be required per deployment due to different number of devices and nodes. Schemas allow developers to express a state layout as a function of system parameters. In practice, developers may implement multiple schemas to iteratively compare representations.

States can also encode bias towards decision horizons. For example, Tesauro et al. describe a choice of state encoding in the context of server resource allocation via a discretized mean request arrival rate \cite{TesauroJongDasEtAl2006}. Their state includes both the current mean arrival rate and the one from the prior observation interval to relate the impact of actions to arrival rate. In workload management tasks, the workload generating process is generally unknown, and future workloads (e.g. request rates or job size) may be independent or correlated to current decisions. State features and preprocessing, e.g. temporal smoothing, must encode such assumptions. In the device placement problem, states are deterministically computed as the graph is traversed in topological order.

In summary, state design for systems-RL is an iterative process which differs from feature design for supervised learning as the state must also capture transition dynamics. To help researchers explore, compare and version state designs, Wield standardizes them through schemas.

\head{Action design.}  Similar to state design, action structure must be designed manually as agent outputs need to be translated to structured system calls, e.g. by generating a special query to update the state of a database. Simple action representations include single binary or categorical decisions where an action selects one of a small number of resources or task slots, e.g. which task to schedule next from a task queue, or which device to assign to an operation on a single node. The term 'action structure' refers to interpreting the outputs of a neural network. For example, in Q-learning, a neural network used to represent the Q-function is designed by creating a  final action selection layer with one neuron per possible integer action. The outputs are interpreted as Q-values. To output multiple distinct actions, multiple action layers may be created. RL practitioners must explicitly consider how decision  problems can be mapped to convenient (i.e. as few distinct actions as possible) representations. 
 
Such representations may not scale to larger problem instances if the number of actions directly corresponds to problem size.  Consider a device placement problem scheduling a large computational graph across a cluster with tens of thousands of devices where each device would then correspond to an action. Without an informative prior, an agent would have to first observe performance dependencies across all types of devices. This would require an impractical number of samples to explore action combinations.

This is in contrast to well-conditioned continuous action spaces (e.g. for a physical actuator) where small changes in output correlate to small and predictable changes in the state trajectory. Large discrete action spaces may require task decomposition. If similarity between actions in large discrete action spaces is known in advance, actions can also be selected in a multi-stage approach whereby first an action in a promising region of the action space is selected, and a nearest-neighbor lookup is subsequently performed to identify a fitting local action \cite{Dulac-Arnold2015}.

Listing \ref{li:task-schema} shows two simplified single-task schemas for the device placement problem. One schema defines a one-dimensional input-array for a recurrent architecture \cite{hierarchical2018} while the other defines layouts for a graph neural network \cite{Addanki2019}. Both share the same action layout based on available input devices. The example illustrates how system-specific configurations are used to define layouts for states and actions.

\begin{lstlisting}[float,caption={Task schemas define programmatic layout based on deployment-specific parameters.},belowskip=-5mm,label=li:task-schema,moredelim={[is][emphstyle]{@@}{@@}}]
NODE_OPTIONS = ['is_current_node', 'is_placed']
MAX_NEIGHBORS = 5

class PlacementSchema(Schema):
  def _build_outputs(self, devices):
    return IntBox(low=0, high=len(devices))

class RecurrentSchema(PlacementSchema):
  def _build_inputs(self, input_graph, devices):
    num_ops = count_ops(input_graph)
    return FloatBox(shape=(num_ops, len(devices)))

class GraphSchema(PlacementSchema):
  def _build_inputs(self, input_graph, devices):
    num_ops = count_ops(input_graph)
    num_options = len(NODE_OPTIONS)
    return Dict({
      'embeddings': FloatBox(
        shape=(num_ops, num_options + len(devices))),
      'current_node_num': IntBox(low=0, high=num_ops),
      'in_neighbors': IntBox(shape=(num_ops, MAX_NEIGHBORS)),
      'out_neighbors': IntBox(shape=(num_ops, MAX_NEIGHBORS))
    })
\end{lstlisting}
In summary, schemas define physical layouts of states and actions, and decouple them from transition dynamics.

\subsection{Converters}
Where schemas correspond to physical layouts, converters are adapters expressing how system metrics, configuration parameters, and query languages or custom protocols correspond to numerical representations within an optimization.

There is a many-to-many relationship between schemas and converters. A schema specifying a layout can be used by different converters, and a converter may work with different schemas. Schemas constrain how decision model is encoded structurally (layout), converters specify how this encoding is achieved from raw system information (content). Listing \ref{li:converter-api} shows the conversion API provided by Wield  with an example implementation for device placement.
\begin{lstlisting}[float,caption={Wield converter API example to translate between agent and system views..},belowskip=-5mm,label=li:converter-api,moredelim={[is][emphstyle]{@@}{@@}}]
# Maps system metrics to state inputs.
def system_to_agent_state(system_state)
  current_op_id = system_state.current_op.id

  embeddings = []
  for op in self.schema.ops:
    is_current_node = op.id == current_op_id
    is_placed = op.id < current_op_id
    one_hot_devices = one_hot(
      op.node.id, len=self.schema.num_devices))
    empbeddings.append(
       (is_current_node, is_placed, one_hot_devices))
  in_neighbors = get_input_neighbors(system_state.current_op)
  out_neighbors = get_output_neighbors(system_state.current_op)
  return (embeddings, current_op_id, 
               in_neighbors, out_neighbors)
  
# Maps system command to numerical representation.
def system_to_agent_action(system_action):
  # System action is device name
  return self.schema.device_name_to_index[system_action]
  
# Maps system metrics to single numerical reward.
def system_to_agent_reward(system_metrics)
  return -system_metrics['run_time']
 
 # Maps agent outputs to system command
def agent_to_system_action(agent_action)
  return self.schema.index_to_device_name[agent_action]
\end{lstlisting}
Workflows invoke the converter API to translate between system and agent representation.

\subsection{Task architectures}\label{Wield-task-architectures}
Schemas and converters help decouple system-specifics from task representation in RL for a single task. Task graphs organize tasks into independent sub- and multi-task architectures. 

\textbf{Shared-parameter tasks} are multi-task architectures where a single end-to-end differentiable architecture has multiple task output networks which each emit separate actions per step. 
\textbf{Independent tasks} are task architectures where separate learners focus on different sub-tasks, e.g. in the case of hierarchical decomposition or parallel independent tasks.

\begin{figure}[t] 
\centering
\includegraphics[scale=0.5]{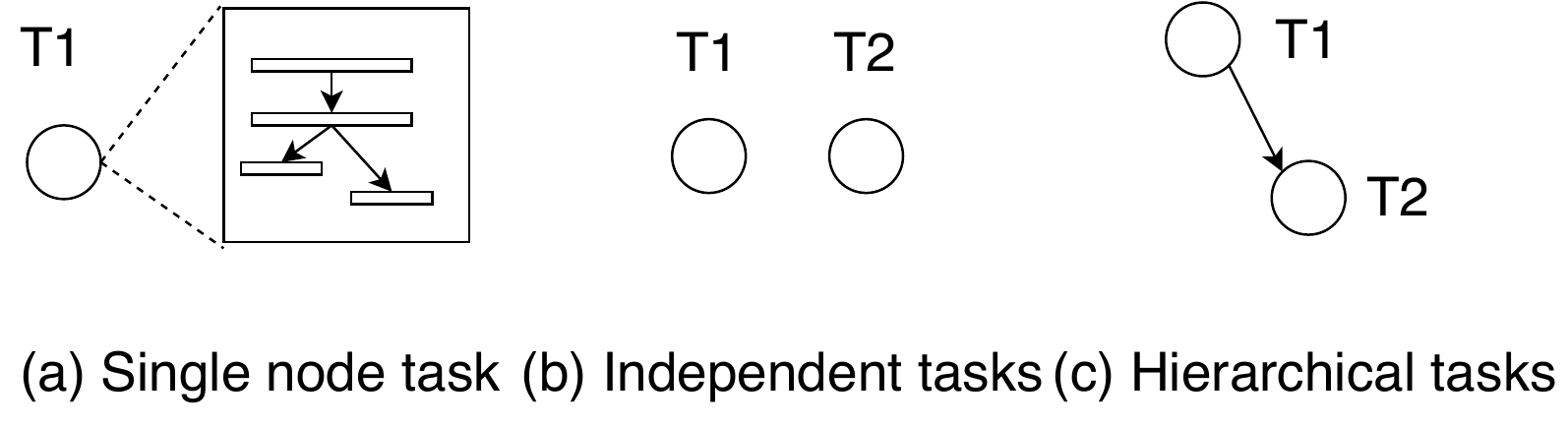}
\caption{Basic task architectures. (a) single task node contains multi-task architecture with shared network (b)  multiple independent learner instances. (c) a hierarchical task dependency.}
\label{fig:task-graphs}
\vspace{-5mm}
\end{figure}

Non-trivial task graphs occur through task decomposition (Figure \ref{fig:task-graphs}). Hierarchical task decomposition refers to tasks organized as directed acyclic graphs where outputs from single tasks (vertices) are used as input states (edges) to other task vertices. Independent tasks refers to a scenario where multiple learners interact with an environment, possibly learning at different time scales. Hierarchical reinforcement learning has been studied in a variety of contexts with the most well known approach being the options framework \cite{Sutton1999}. There, a top-level policy chooses between different sub-policies (options) to execute over a time-frame (until the sub-task terminates). In Wield, we focus on work flows where users manually identify task hierarchies as a means of encoding domain knowledge.

Hierarchical designs to organize resources at different granularities are also a core element of systems research (e.g. cache hierarchies, hierarchical scheduling). However, hierarchical RL has found limited attention in the systems community as a means to manage large state and action spaces. This could be due to most open source implementations focusing on single-agent scenarios or unstructured collections of policies (e.g. RLlib \cite{Liang2018}).

Task graphs in Wield simplify factorization of tasks into different sub-tasks which may train and act jointly, or at different time-scales. Task objects primarily encapsulate distinct agents or any other optimization implementing OpenAI gym \cite{openaigym} interfaces. Hierarchical tasks often require to transform the output of one task before inputting it to a subsequent task,  e.g. by enriching it with additional environment information or preparing a specific input format. Nodes in a task graph hence further encapsulate pre-and post-processing for each  sub-task. Edges in the graph are implicitly created by creating one task as a sub-task  of another task in the same task-graph. When performing inference, task outputs are routed through the task graph based on user-defined directed edges between tasks, and the results of all tasks during execution are returned.

In this section, we introduced a light-weight set of software primitives for modularized task design. Next, we present progressive randomization as the guidance mechanism to evaluate representations.
\begin{table*}[h]
  \caption{Progressive randomization protocol overview. Each class specifies a different level of non-determinism.}
  \label{progressive-randomization}
  \centering
  \begin{tabular}{llll}
    \cmidrule{1-4}
   Class     & Optimization seed & Workload randomization & Example use cases\\
    \midrule
   $ \mathbf{C_0}$ &  Fixed  & Fixed blackbox task   & Iterate and debug representation\\
   \midrule
   $ \mathbf{C_1}$ &  Random  	&  Fixed blackbox task   & Weight initialization sensitivity  \\
   \midrule                          
   $ \mathbf{C_2}$&  Random &  Randomized blackbox task   & Model sensitivity to task parameters \\
   \midrule
   $ \mathbf{C_3}$ &  Random &  Fixed in-distribution generalization   & Understand sample requirements  \\    
   \midrule
   $ \mathbf{C_4}$ &  Random&  Randomized in-distribution generalization   & Customized production use \\
   \midrule
   $ \mathbf{C_5}$ &  Random &  Fixed out-of-distribution generalization   & Robustness against unforeseen inputs  \\
   \midrule                                                                        
   $ \mathbf{C_6}$ &  Random &  Randomized out-of-distribution generalization   & Production use without customization  \\
    \bottomrule
  \end{tabular}
\end{table*}
\section{Progressive randomization}
\subsection{The case for task randomization}
A key obstacle when assessing model capabilities is the use of fixed workloads. In domains like database management, query processing benchmarks often focus on narrow application scenarios with small query sets (e.g. TPC-H, TPC-C). Leis et al. proposed the Join Order Benchmark which contains 113 queries specifically designed to investigate join estimation capabilities in query optimisers \cite{Leis2015}. In device placement, researchers have relied on fixed graphs of standard architectures which may differ in implementations \cite{hierarchical2018}, custom variants of common architectures \cite{Addanki2019}, or proprietary datasets \cite{Paliwal2019}. 

While hand-designed workloads can highlight particular weaknesses or strengths of a system, they nevertheless are prone to over-fitting small test sets. We argue that the design of RL mechanisms for systems can benefit from synthetic workload mechanisms with configurable task difficulty as a means to understand both training and inference behaviour. 

Reasoning about non-determinism when evaluating stochastic optimization mechanisms requires  distinguishing deterministic and non-deterministic elements in both workload and optimization procedure (network weight initialization). In Wield, we construct workloads from the perspective of changing between several evaluation and randomization modes. We distinguish between blackbox and generalization mode. In blackbox mode, a single workload instance (e.g. a single set of queries or jobs) is generated and a model is trained and evaluated on that same instance. In generalization mode, training is executed on different instances than the ones used in the final evaluation.

Both modes can be executed with varying levels of randomization. Workload determinism refers to deterministic behaviour of task instances. Training determinism refers to deterministic initialization and sampling during training. For example, in blackbox mode the generation of the single task instance and the training initialization can both be deterministic. Similarly, in generalization, both the instances used during training and the final test instances can be randomly generated or held fixed. This invites problematic practices such as cherry-picking and presenting only successful combinations of workloads and weight initialization values (while omitting this selection).

In the RL literature, all combinations of blackbox and generalization modes can be found. Comparing results is difficult if authors do not to report which workload elements are held fixed or are subject to randomization, or why a particular sample was chosen. 

\begin{table*}[t]
  \caption{Progressive randomization protocol overview. Each class specifies a different level of non-determinism.  If a range is given without an approximate estimate, this refers to different data sets being  reported at different sample sizes. A * refers to researchers reporting median or mean across random optimization seeds.}
  \label{progressive-randomization-table}
  \centering
  \begin{tabular}{lll}
    \cmidrule{1-3}
   Work     & Objective & Highest $C_i $ reported\\
      \midrule
Neural packet classification \cite{Liang2019} & Classification time/memory & $C_0(n=10^7,s=?,f=?)$ \\
Device placement \cite{hierarchical2018} & SGD iteration time & $C_1(n=10^3,s=?,f=?)$  \\
Device placement \cite{mirhoseini2017device} & SGD iteration time & $C_1(n=2 \times 10^5,s=4,f=1.0)^*$  \\
Join order \cite{Marcus2018} & Query execution time & $C_3(n\approx 10^4,s=?,f=?)$  \\ 
Device placement \cite{Addanki2019} & SGD iteration time & $C_3(n=10^3-10^5,s=?,f=?)$  \\
Cardinality predictions \cite{Ortiz2018} & Prediction error  & $C_3(n\approx10^5-10^6,s=?, f=?)$\\
Language to program \cite{Guu2017} & Program generation & $C_3(n\approx 10^4, s=5,f=1.0)^*$ \\
Spark scheduling \cite{Mao2018b} & Spark job completion times & $C_4(n=10^8,s=?,f=?)$  \\
Congestion control \cite{Jay2019} & Throughput, latency & $C_4(n\approx10^6-10^7,s=?,f=?)$\\
Query Optimizer \cite{2019arXiv190403711M} & Improve query latency & $C_5(n\approx 10^4-10^6,s=50,f=1.0)^*$  \\
Computation graph rewriting \cite{Paliwal2019} & Memory usage & $C_5(n=4 \times 10^5,s=20,f=?)$\\
AlphaZero \cite{Silver1140} & Win game of chess & $C_6(n\approx 10^9,s=?,f=?)$  \\

    \bottomrule
  \end{tabular}
\end{table*}

\subsection{Progressive randomization}
\head{Overview.} Progressive randomization is based on the observation that different randomization modes can serve different phases of design. For example, holding a workload fixed to study robustness against random initialization is valuable when a designer is uncertain if a model design can solve a task at all. Conversely, using a fixed optimization can be useful to study the impact of workload parameters on optimization outcomes. Evaluation difficulties are not inherent to a specific mode of randomization or evaluation. They arise when conflating sources of performance variation or misinterpreting model capabilities.

In supervised learning, projects such as DAWNBench \cite{Coleman2018} have suggested metrics like time-to-accuracy to compare model designs and hardware choices to understand trade-offs in deep learning systems. In contrast, shared RL tasks such as the Malmo Minecraft challenge \cite{Johnson2016} or Unity agents \cite{Juliani2018a} are focused on performance in simulated worlds where randomization is \textit{incidental}. That is, workloads may include some degree of randomization and generalization but these are not varied  to analyse their contribution to agent performance (or lack thereof). Task variation in these scenarios is further constrained by experimental cost. The purpose of randomization is to evaluate  model robustness to both subtle and fundamental changes in workloads. For example, in systems-RL, this requires gathering evidence about plausible workload distributions a controller may encounter. 

Table \ref{progressive-randomization} lists the different evaluation modes in the protocol and their purpose. It also lists example applications. Fixed optimization parameters in practice refer to the random weight initialization strategies in neural networks, and further to the random seed used when sampling mini-batches for stochastic gradient descent as well as policy decisions. Fixed blackbox refers to always training on the same workload or problem instance, while fixed generalization refers to an unseen but fixed test task. Randomized generalization implies that for each reported experiment result, a new test instance was generated. 

Not all possible combinations of non-determinism are present in the protocol. Fixed optimization parameters on fixed workloads are initially useful to produce repeatable results and debug non-optimization components of a task ($C_0$). For subsequent evaluation concerns, they should be randomized to avoid cherry-picking 'lucky' seeds. In $C_1$ and $C_2$, weight initialization and workload instance (e.g. a set of jobs sampled from a workload distribution) are  incrementally randomized.

\head{Generalization.} Subsequent levels evaluate performance on unseen problem  instances. \textit{In-distribution generalization} refers to workload assumptions where the test task is taken from the same distribution training tasks were generated from. For example, the device placement problem can be randomized by varying batch size and sequence unroll lengths on the same architecture  \cite{Addanki2019} (in-distribution), or by testing on entirely new architectures (out-of-distribution). Generalization semantics are complicated by task-specific concerns. For fixed or randomized generalization tasks, there may be no useful measures of how different test tasks are from training examples. Parts or the entire test task could be seen during training, unless the test task is held out and rejected.

Nonetheless, the description of a model to e.g. be in $C_3$ for a certain task gives useful indication of expected behaviour. We refer to being in a class as to meeting application-specific performance objectives under the  given randomization assumptions. For example, a model in $C_2$ which meets randomized blackbox objectives can be used as a direct search tool in practice without requiring to retune hyper-parameters, whereas a model in $C_1$  tuned for a fixed blackbox objective is customized to a single deployment or task context. Distinguishing model classes sets expectations and allows researchers to effectively communicate evaluation designs.

Generalization concerns in deep reinforcement learning are poorly understood. They are not well captured analytically but rather empirically per task. This is primarily a consequence  of limited understanding on generalization capabilities of neural networks as policy vehicles. A model may be in different classes depending on the number the samples is trained on. In particular, researchers at OpenAI highlighted in their work on competitive DOTA that massively increasing model and sample  size can induce qualitatively different generalization behaviour \cite{openaidota}. Moreover, even for the same hyper-parameters, a large fraction of random weight initialization and optimization seeds may vary drastically in performance \cite{Henderson2017}. 

We propose to describe models based on the these empirical properties. Progressive randomization classification thus includes (i)  the number of state transitions experienced during training $\mathbf{n}$,  (ii)  the number of random seeds used for weight initialization and optimizations  $\mathbf{s_r}$,
and  (iii)  the observed frequency $f$ where learning objectives were achieved.

For example, a model may be described as $C_1(n=10^7, s=10, f=0.4)$ to communicate empirical success when training 10 million samples and trying 10 different random seeds, where 4 of 10  trials met the objective. In the following notation, we omit $s$ and $f$ from notation when only discussing sample count or class membership. Communicating success rates is especially important when considering the training cost on single-tasks without generalization.

As sample collection cost varies drastically between tasks, conditioning class membership on sample size is useful for estimates on model transfer on tasks with different sample collection cost. For  example, the same model may be in $C_1(n=10^4)$ but in $C_3(n=10^6)$ as robustness to inputs increases with experiences  seen during training. The number of sample trajectories seen during training may also correlate with the observed frequency of reaching an objective.

\head{Limitations.} Progressive randomization encourages shared understanding of model capabilities across problem domains. Practitioners can use it to incrementally test new implementations. Several dimensions regarding model scale, the cost of featurization, and other hidden cost are not captured. The protocol also does not replace standard considerations on experiment design or statistical analysis. The classification system is intentionally simple to serve as a low-overhead summary of design assumptions. While only a starting point, progressive randomization constitutes the first explicit evaluation protocol for focused on delineating workload randomization.

\subsection{Prior work viewed through progressive randomization}
We use progressive randomization as a lens on prior work in research and applied RL. Table \ref{progressive-randomization-table} classifies selected prior work, sorted by class membership. The classification immediately illustrates problem progress. For example, in the device placement problem, Mirhoseini et al.'s initial work \cite{mirhoseini2017device} with manual operation grouping required orders of magnitudes more samples than their subsequent work using a hierarchical approach \cite{hierarchical2018}. Both operated in a fixed blackbox setting. Addanki's et al.'s \cite{Addanki2019} and Paliwal et al's recent work \cite{Paliwal2019} utilizing graph neural networks then illustrates progress towards generalization through permutation-invariant representations. 

In our survey, we found subtle differences in evaluation randomization which can be made make explicit through progressive randomization. For example, Addanki et al. generate random variations of computation graphs for training and testing, but both sets are fixed ($C_3$).

A similar progress pattern can be observed in database tasks. In their first work on join order enumeration, Marcus et  al. \cite{Marcus2018} used a policy optimization method on a fixed set of training and test queries, the Join Order Benchmark (JOB) \cite{Leis2015}. Training with randomized optimization parameters yields $C_3(n\approx 10^5)$. In subsequent work, Marcus et al. proposed a learned query optimiser \cite{2019arXiv190403711M} which they evaluate on several tasks, including a fixed set of out-of-distribution queries $C_5(n\approx 10^4-10^6,s=50,f=1.0)$. We also observe that training workloads in database applications were often generated by augmenting fixed existing query sets (TPC-H, IMDB). It would be desirable for the systems-RL community to develop shared standards on training and test randomization. 

Many approaches do not report explicitly how workload randomization and optimization parameters were selected which makes classification difficult. If a fixed task is presented without reporting number of training trials, seeds, or randomization assumptions (i.e. a potentially cherry picked single random seed), we assume $C_0$. 

Few of the applied works we surveyed explicitly report on failure modes, despite often using appendices to communicate other training hyper-parameters. This highlights the need for more explicit evaluation protocols. Evaluation times for systems-RL can vary drastically between microseconds and minutes. Sample-size classification helps researchers evaluate if a simulator may be needed.

Finally, prior systems  applications of RL are not typically defined through a binary objective such as winning a game or reaching a score threshold. Performance objectives are explorative, e.g. outperforming problem-specific baselines. This can obfuscate practical utility without cost-benefit analysis on implementation cost.
\begin{figure}[ht]
\centering
\begin{subfigure}[t]{.4\textwidth}
\includegraphics[width=\textwidth]{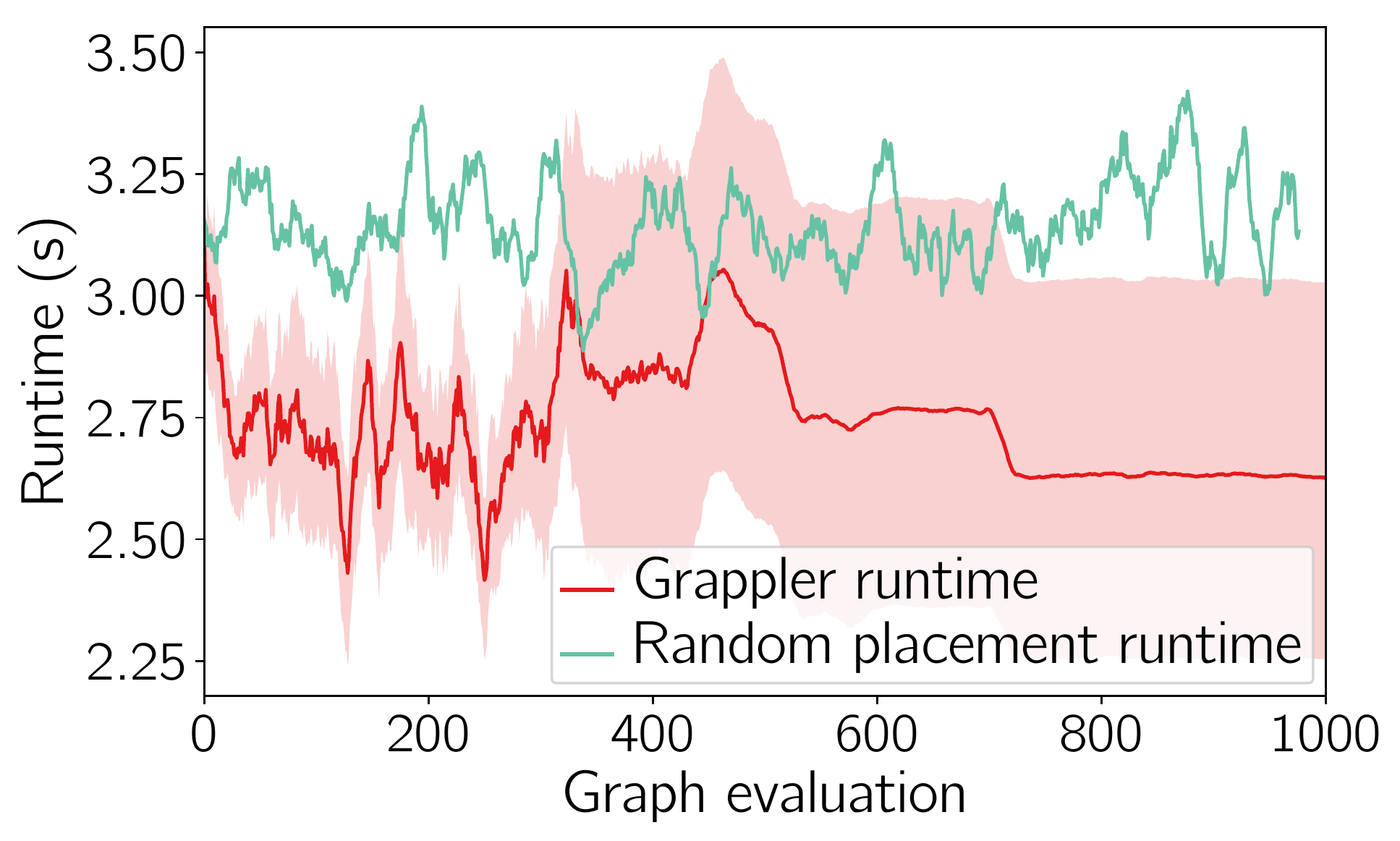}
\end{subfigure}
\begin{subfigure}[t]{.4\textwidth}
\includegraphics[width=\textwidth]{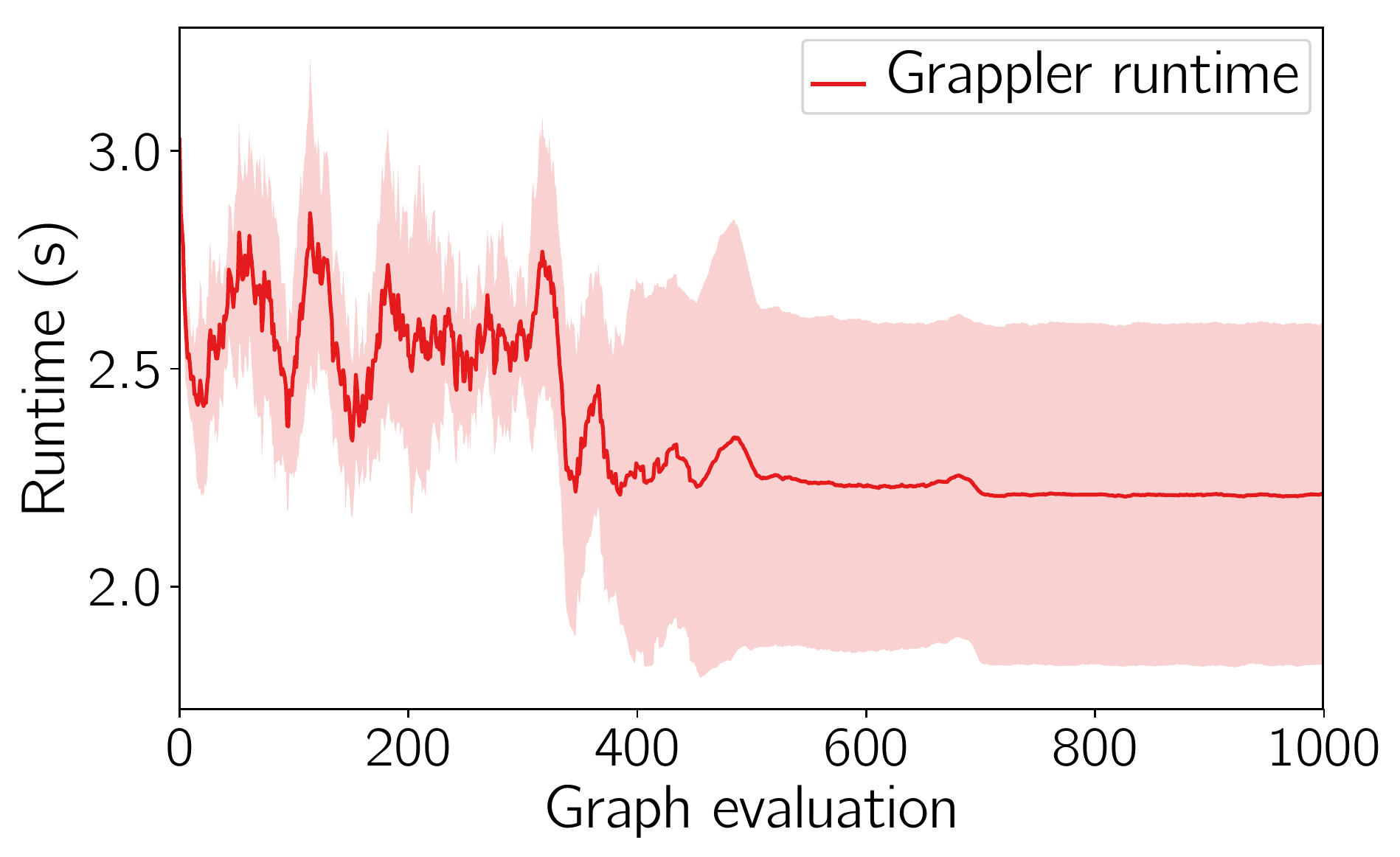}
\end{subfigure}
\caption{Reproducing the hierarchical placer on the fixed NMT graph. Top: Fixed seed. Bottom: Random seed sampled per trial. \label{fig:grappler-blackbox}}
\vspace{-2mm}
\end{figure}
\section{Evaluation}
We illustrate systematic assessment on the device placement problem. We then subsequently implement a competing model and again incrementally evaluate it.

\subsection{Reproducing prior work}
We use progressive randomization to evaluate the open source TensorFlow device placer ('Grappler') available as part of the grappler module\footnote{\url{https://github.com/tensorflow/tensorflow/tree/master/tensorflow/python/grappler}} \cite{hierarchical2018}. 

\head{Setup.} As a benchmark, we use the neural machine translation (NMT) architecture evaluated by all prior device placement  work. While some models were trained in a distributed setting, reported NMT results (\cite{hierarchical2018}, Figure 3) illustrate strong performance improvements within hours of training using a single node. NMT is an attractive benchmark task because the variants tested consist of an encoder-decoder architecture with multiple LSTM layers. Prior work shows placements must split training batches non-trivially across GPUs and time dimensions.

The open source placer was not able to run evaluations on Google's NMT implementation with its own evaluation utilities. Serialization of a number of related metagraph components failed. The results presented here were obtained by directly instantiating the TF graph and calling training operations. This significantly increases cost per measurement. Each measurement was given one warm-up run, and the subsequent measurement was reported to the controller. The open source hierarchical placer decays its learning rate to 0 within 1000 updates, corresponding to the results reported in the paper. Here, all results were ran at least 1000 steps.

\head{Fixed blackbox.}
We begin with the fixed random seed, fixed workload setting ($C_0$). Figure \ref{fig:grappler-blackbox} (top) illustrates runtime of the training operation (i.e. a single iteration of mini-batch stochastic  gradient descent). Results are averaged across runs, shaded areas indicate 1 standard deviation confidence intervals. We used the random seed supplied by the default configuration in the open source implementation (1234), and repeated the experiment 10 times. 

The placer identified improved placements in most runs with a mean final improvement (measured as the mean of the final 10 steps against the initial runtime) of 52\%. One run failed to substantially improve in the end (5\%). Invalid placements were removed from the figures (assigned runtime value 100 in the implementation).

We break down both the \textit{final} relative improvements and the best-seen solution during training in Table \ref{placer-fixed-improvement-table}. Results show that i) all trials identified significant improvements during training, and ii) some trials diverged.
\begin{table}[t]
  \caption{Fixed blackbox improvements found by Grappler.}
  \label{placer-fixed-improvement-table}
  \centering
  \begin{tabular}{lll}
    \cmidrule{1-3}
   Trial     & Final model &  Best seen \\
    \midrule
Trial 1 & 57.0\% & 69.0\% \\
Trial 2 & 67.0\% & 71.0\% \\
Trial 3 & 25.0\% & 69.0\% \\
Trial 4 & 51.0\% & 73.0\% \\
Trial 5 & 5.0\% & 74.0\% \\
Trial 6 & 70.0\% & 73.0\% \\
Trial 7 & 69.0\% & 69.0\% \\
Trial 8 & 47.0\% & 73.0\% \\
Trial 9 & 68.0\% & 70.0\% \\
Trial 10 & 66.0\% & 69.0\% \\
  \end{tabular}
  \vspace{-5mm}
\end{table}
Divergence even occurs with a fixed random seed and a fixed workload, likely due to noise in the reward. We also contrast learned values against entirely random placements and groupings in Figure \ref{fig:grappler-blackbox} (top) which fail to find good placements.

In Figure \ref{fig:grappler-blackbox} (bottom), the same graph is evaluated on 10 randomly chosen seeds (fixed workload, randomised optimization parameters, $C_1$). Mean improvement was 72\% with all random seeds achieving over 70\% improvement. The fixed seed led to higher variance than the random seeds.

While restricted to a single graph, reproduced results confirm the paper's claims of reliable improving runtimes in black-box settings. With a success criterion of at least $30\%$ improvement (minimum improvement in paper $35.9\%$), we report $C_1(n=1000, s=10, f=1.0)$.

\begin{figure}[ht]
\centering
\begin{subfigure}[t]{.4\textwidth}
\includegraphics[width=\textwidth]{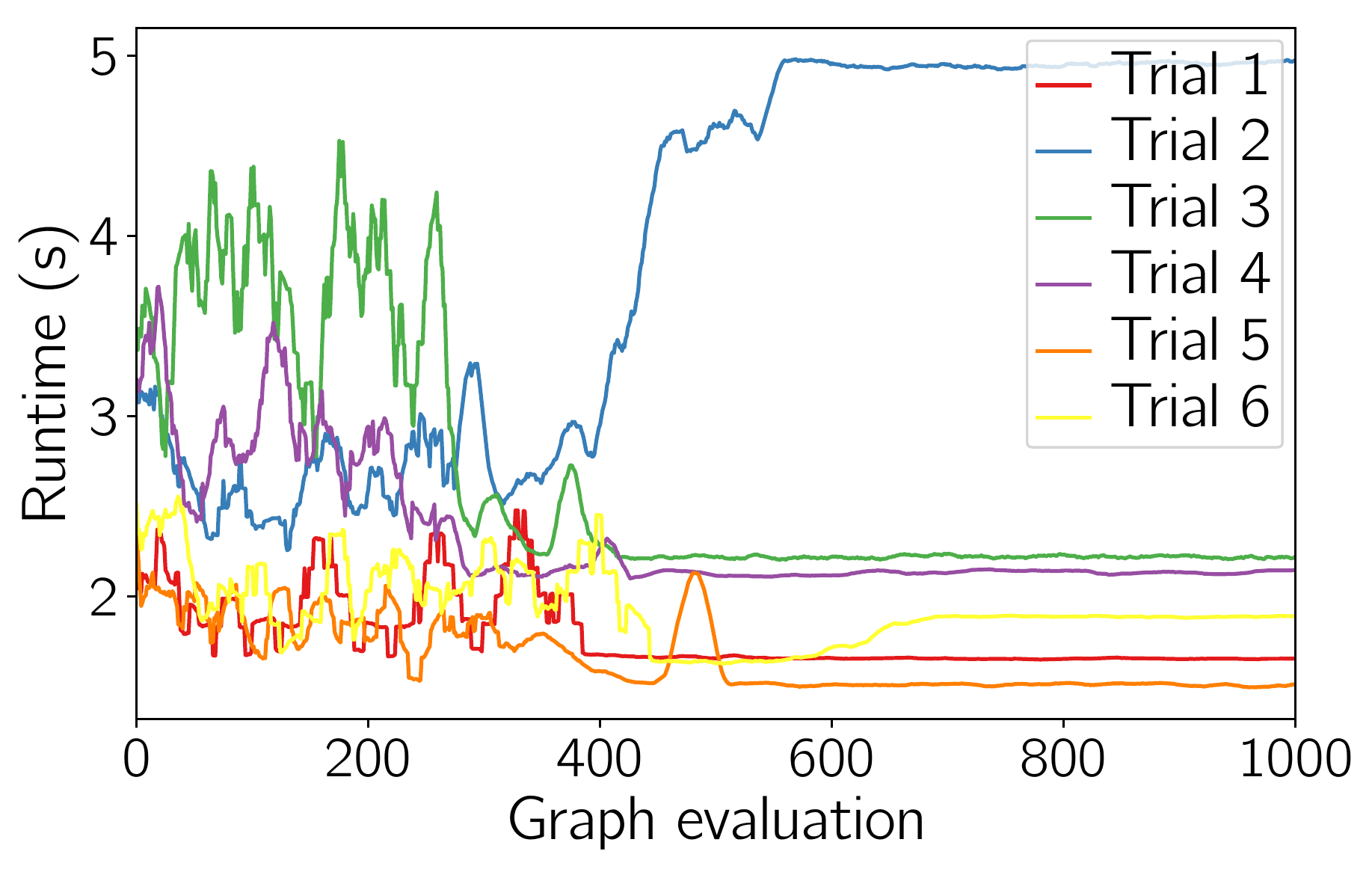}
\end{subfigure}
\begin{subfigure}[t]{.4\textwidth}
\includegraphics[width=\textwidth]{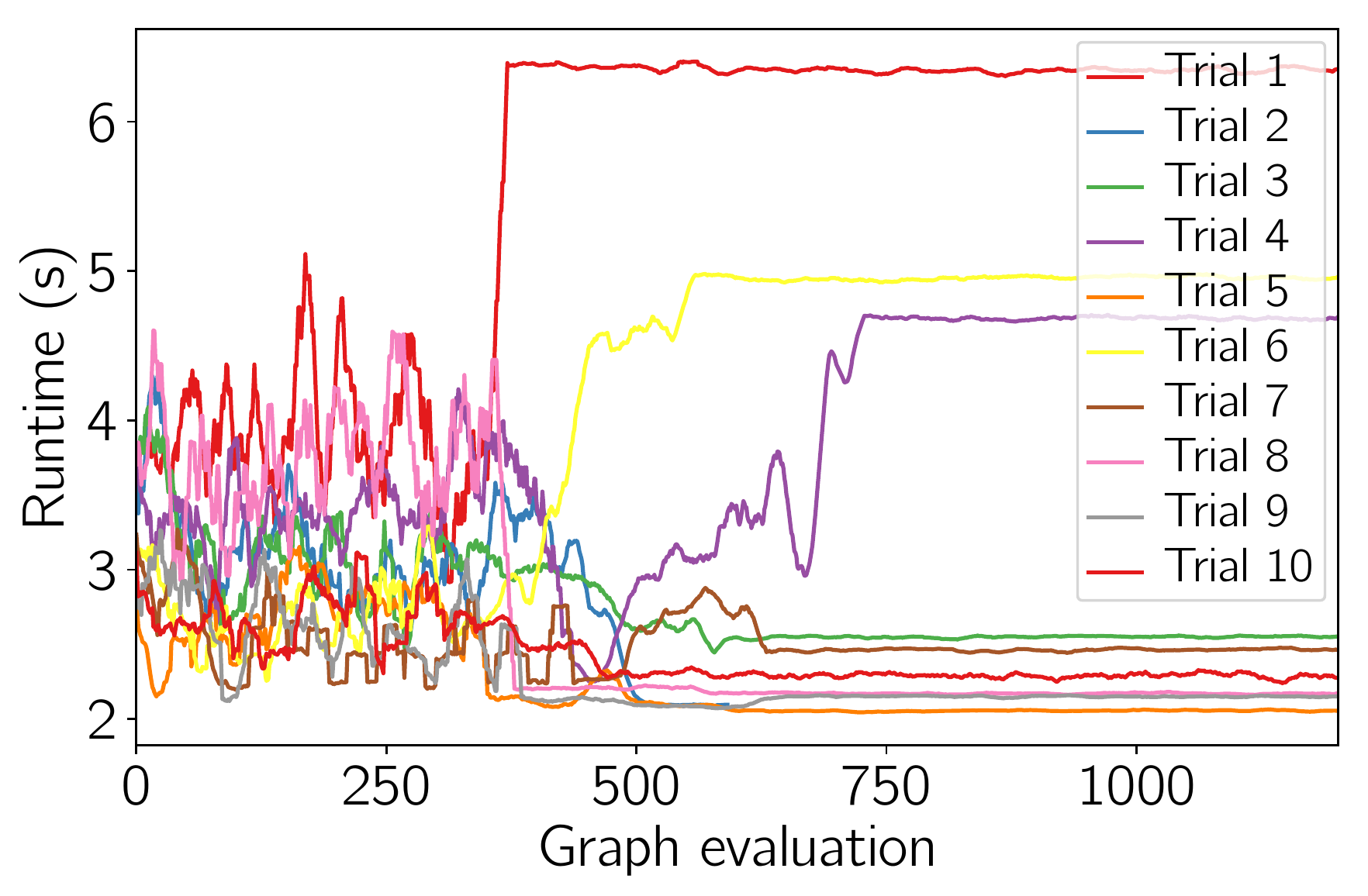}
\end{subfigure}
\caption{Randomized blackbox evaluation. Top: Random blackbox, random seed.  Bottom: Repeats of failed trial 2. \label{fig:random-blackbox-random-seed}}
\vspace{-2mm}
\end{figure}

\head{Randomized blackbox.}
In Figure \ref{fig:random-blackbox-random-seed}  (top), we modify task parameters by varying batch size and unroll lengths in the recurrent network to create a randomized blackbox setting ($C_2$) where both graph and optimization seed are varied. Best runtimes are expected to differ due to different graph sizes. Of six trials, three succeeded, one failed entirely (1), and one (5) diverged from an effective configuration ($C_2(n=1000,s=6, f=0.83)$).

Did trial 2 fail due to posing a more difficult placement task, or due to random initialization?  In case of unclear failures, we decrease randomization levels and re-evaluate the failed task. We repeated the failed trial as a fixed blackbox task with randomized optimization parameters. Figure \ref{fig:random-blackbox-random-seed} (bottom) shows results of rerunning the same task nine more times for a total of ten trials. Results include more fails and diverged results but also succeeding runs. Three trials failed to improve placements significantly. Performance variation is further higher than in the published result, as graph variations affect failure rate.

\begin{figure}[ht]
\centering
\begin{subfigure}[t]{.235\textwidth}
\includegraphics[width=\textwidth]{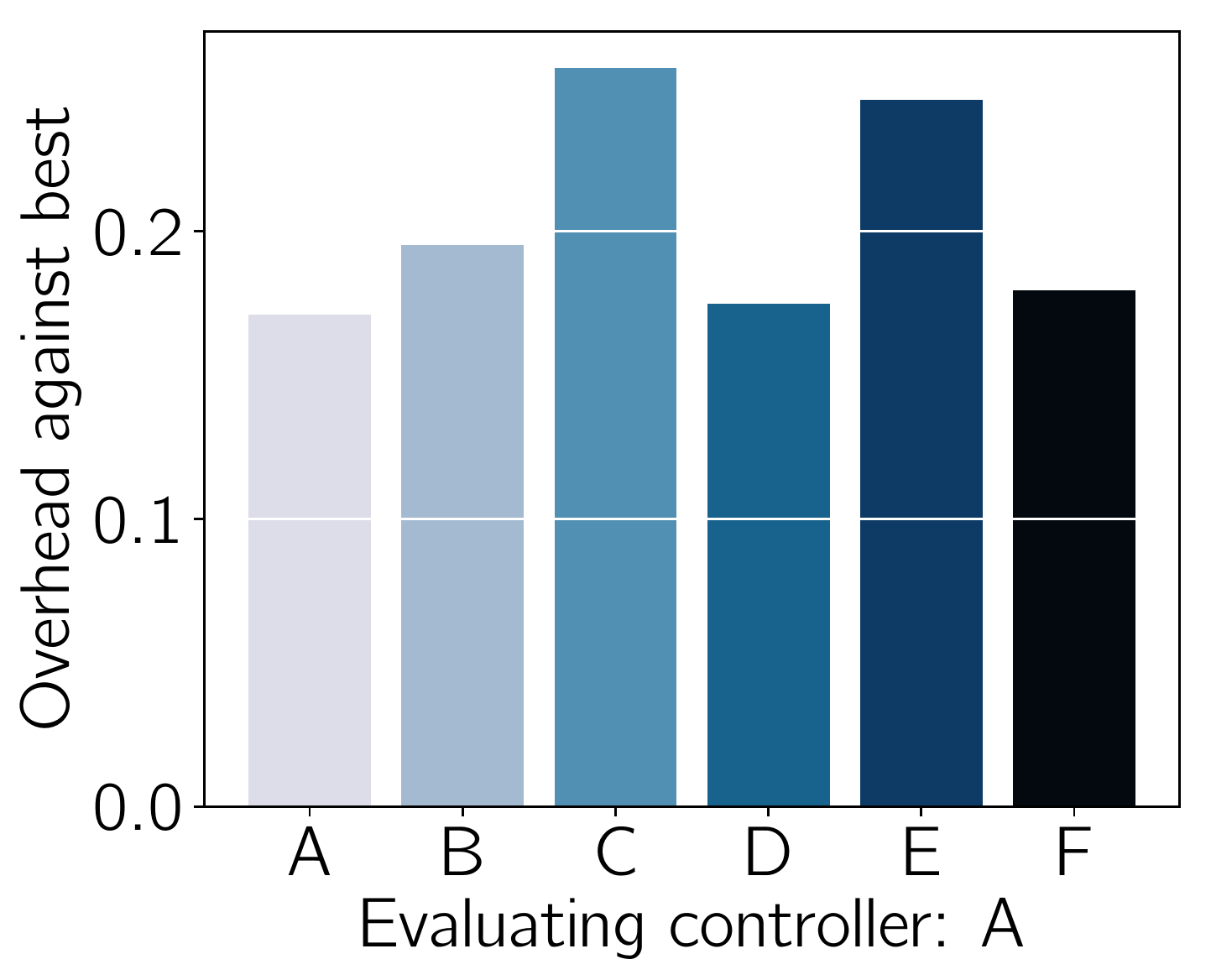}
\end{subfigure}
\begin{subfigure}[t]{.235\textwidth}
\includegraphics[width=\textwidth]{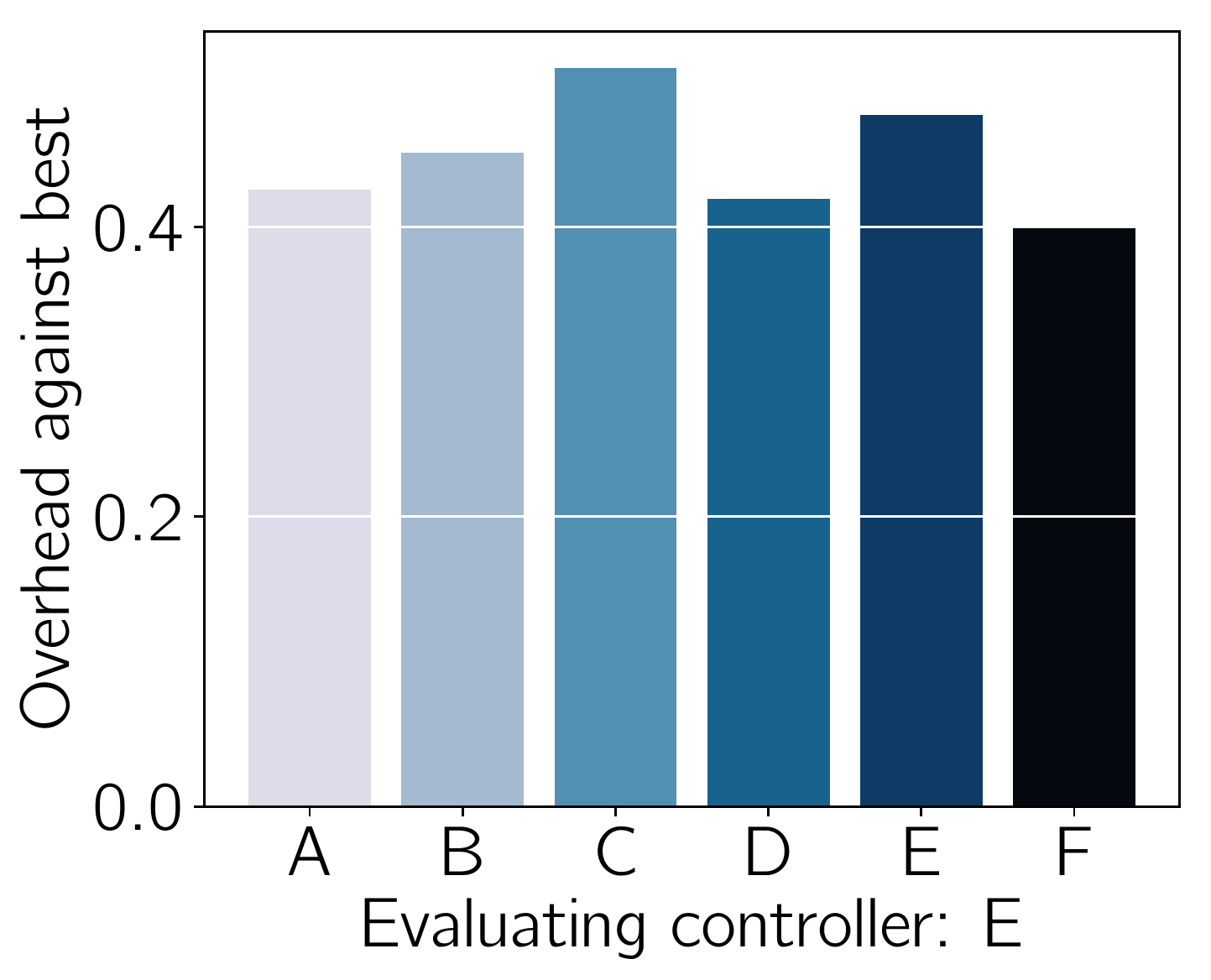}
\end{subfigure}
\caption{Generalization overhead example. \label{fig:generalisation-example}}
\vspace{-5mm}
\end{figure}

\head{Fixed in-distribution generalization.}
Next, we consider generalization capabilities. We first evaluate the final trained model in each trained randomized graph against all other trials' graphs. Figure \ref{fig:generalisation-example} show two examples of cross-graph comparisons. Final models perform significantly worse against the best seen solutions when training specifically on the respective graph, with overheads against the best solution ranging between 20\% - 50\%. Analysis of placements shows (i) during training on a particular graph, non-trivial placements using  all devices are identified during exploration, and (ii) diverged final placements default to single-device (CPU only) or single-GPU. When evaluating generalization, placements for slightly varied graphs frequently defaulted to trivial single-GPU decisions.

The failure to identify non-trivial placements for varied graphs is hence both a result of models diverging and limited model capability. We compare detailed generalization results and classify against our placer in the next section. 

\subsection{Implementing a placer with Wield}
Next, we use Wield's primitives to implement and evaluate a new placer. Our implementation combines insights from the open source hierarchical placer and Addanki's recent work on using graph neural networks \cite{Addanki2019}. Addanki et al. rely on manual grouping and identify effective generalizable placements based on i) incremental evaluation of placements, and ii) a neighborhood embedding in a graph neural network based on computing parallel, parent, and child-operation groups.

Our goal is to investigate if Wield's abstractions can be used to re-implement customized architectures. We loosely combine both approaches using Wield's task graph abstraction to implement an hierarchical graph network placer. Grouper and placer are proximal policy optimization agents \cite{Schulman2017} learning independently at different time scales. 

We repeat the same NMT experiments used on the hierarchical  placer. To calibrate hyper-parameters, we ran five trial experiments to identify effective learning rate schedules and incremental reward modes before final experiments. Our main finding was that the grouper required a more aggressive learning rate schedule, because initially random groupings caused the neighborhood computation in the placer to diverge (all groups were connected to all others).
\begin{figure}[t] 
\centering
  \includegraphics[scale=0.4]{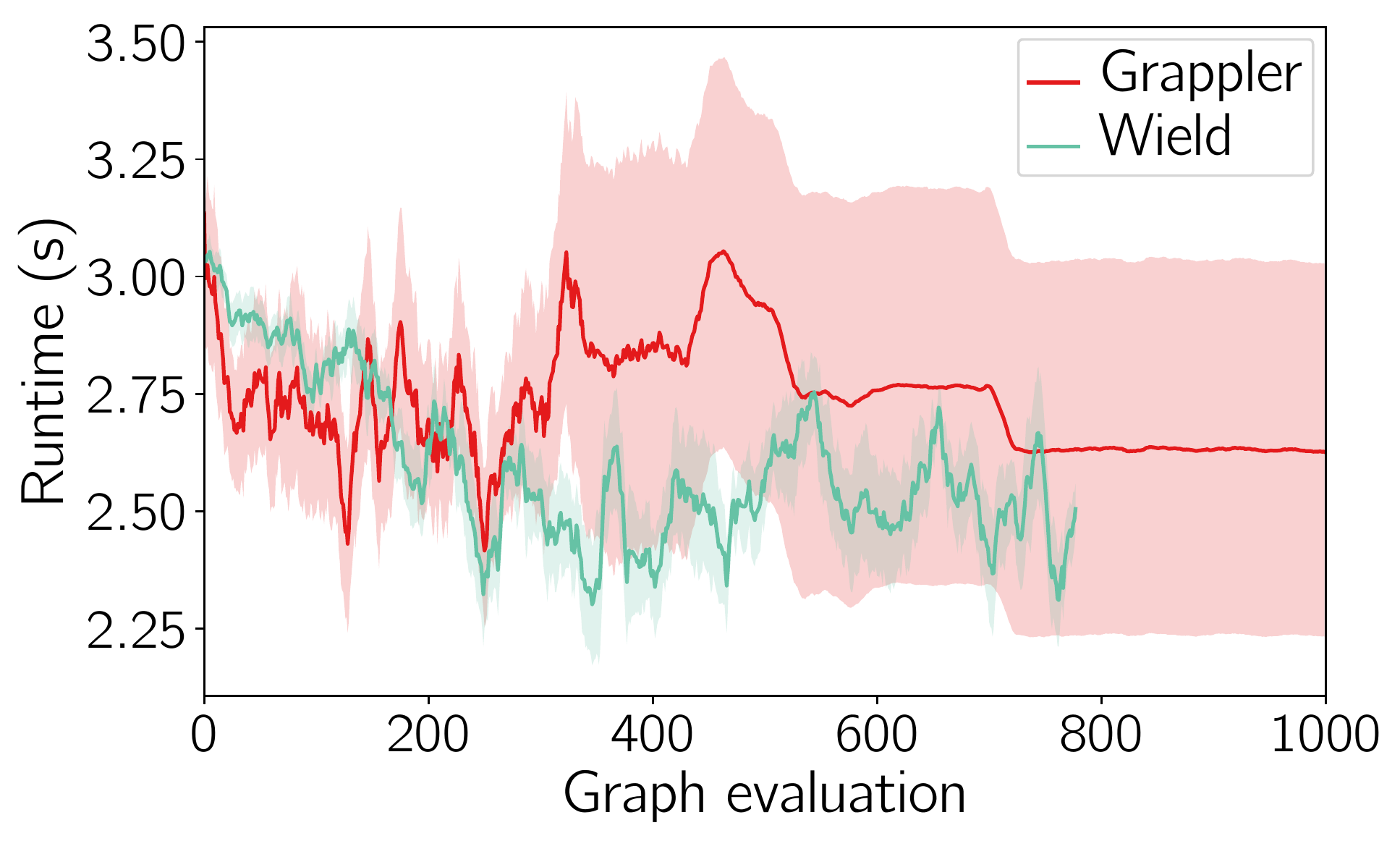}\caption{Open source hierarchical placer versus Wield placer.}
\label{fig:wield-grappler-fixed}
\vspace{-5mm}
\end{figure} 

Figure \ref{fig:wield-grappler-fixed} compares operation runtimes over the course of training. Due to  cost, we terminated some experiments early after no more improvement was observed. Both implementations frequently found their best placements within 500 graph evaluations. We repeated the randomized blackbox experiment with 6 new graphs ($C_2$) to evaluate sensitivity to graph variations. In table \ref{randomized-blackbox-wield-grappler-table}, we compare relative runtime improvements across trials between Grappler and Wield. Mean improvements are similar and differences not statistically significant (for $p<0.05$). 

Finally, we also repeated the cross-graph generalization experiment to investigate generalization capabilities of the structured neural network representation. Since the network computes a permutation invariant embedding of operation neighborhood, higher robustness to input variants should be observed. We show a detailed breakdown of both approaches' generalization capability by showing how the  final model trained on a graph (rows) performs in terms of relative runtime improvement on another graph (columns) (Tables \ref{grappler-generalisation} and \ref{wield-generalisation}).
\begin{table}[t]
  \caption{Relative improvements in randomized blackbox scenario.}
  \label{randomized-blackbox-wield-grappler-table}
  \centering
  \begin{tabular}{lll}
    \cmidrule{1-3}
   Trial     & Grappler best & Wield best \\
    \midrule
Trial 1 & 22.6\% & 37.4\% \\
Trial 2 & 37.4\% & 38.2\%\\
Trial 3 & 37.8\%  & 32\% \\
Trial 4 & 36.7\%  & 37.4\%\\
Trial 5 & 37.2\%  & 41.5\% \\
Trial 6 & 34.5\%  &  48\% \\
Mean & 34.4\% (5.4\%) & 39.1\% (4.9\%) \\
  \end{tabular}
  \vspace{-5mm}
\end{table}
\begin{table}[!htb]
    \caption{Cross graph generalization breakdown of Grappler models. \label{grappler-generalisation}}
      \centering
 \begin{tabular}{lrrrrrr}
    \cmidrule{1-7}
 On graph  & A & B & C & D & E & F \\
    \midrule
Model A & 0.31 & 0.35 & 0.28 & 0.31 & 0.27 & 0.23  \\
Model B & 0.31 & 0.37 & 0.29 & 0.32 & 0.27 & 0.26  \\
Model C & 0.25 & 0.30 & 0.23 & 0.26 & 0.22 & 0.18  \\
Model D & 0.10 & 0.15 & 0.06 & 0.10 & 0.06 & 0.03  \\
Model E & 0.16 & 0.21 & 0.14 & 0.16 & 0.14 & 0.09  \\
Model F & 0.28 & 0.32 & 0.26 & 0.29 & 0.25 & 0.21  \\
  \end{tabular}
\end{table}
\begin{table}[!htb]
    \caption{Cross graph generalization breakdown of Wield models. \label{wield-generalisation}}
      \centering
 \begin{tabular}{lrrrrrr}
    \cmidrule{1-7}
 On graph  & A & B & C & D & E & F \\
    \midrule
Model A & 0.21 & 0.60 & 0.54 & 0.50 & 0.39 & 0.19  \\
Model B & -0.30 & 0.33 & 0.24 & 0.09 & -0.02 & -0.22  \\
Model C & -0.08 & 0.44 & 0.36 & 0.23 & 0.15 & 0.05  \\
Model D & 0.09 & 0.34 & 0.32 & 0.16 & 0.05 & -0.08  \\
Model E & -0.04 & 0.45 & 0.42 & -0.49 & -0.62 & -0.83  \\
Model F & 0.39 & 0.42 & 0.65 & 0.58 & 0.53 & 0.44  \\
  \end{tabular}
\end{table}
Grappler's placer only in few instances significantly improved the initial placement ($> 30\%$), with a classification of  $C_4(n=1000,s=36,  f=0.22)$. Wield's placer achieves $C_4(n=1000,s=36,  f=0.47)$ and exhibits successful in-distribution generalization with model F. Overall, the Wield placer performs like the custom-built tuned Grappler on blackbox tasks, and indicates potential on generalization. We stress that generalization training would normally be executed across a distribution of tasks (as Addanki et al. do), whereas we (due to significant cost) only trained on one single graph. Due to limited generalization success, we did not evaluate higher randomization levels.

\subsection{Discussion}
We systematically evaluated the open source placer through progressive randomization.  The hierarchical placer with high frequency identified effective placements in a randomised blackbox scenario ($C_2$) but failed to generalize even to slight input variations ($C_4$). Our experiments highlight the significant cost associated with evaluating a model on real-world system, even with a full set of pre-tuned hyper-parameters. Including all calibrations of the custom evaluation due to bugs in the open source code, it cost us $\$5000$ to assess the hierarchical placer on public cloud infrastructure. 

We also showed that using Wield, a competitive placer could be implemented by combining off-the-shelf algorithmic components. In both placers, models diverged after identifying effective placements, and learning rate schedules would need to be tuned to high precision to prevent this. Both the true cost of such calibrations and the impact of workload randomization have not been widely discussed in the systems-RL community. With Wield and progressive randomization, we aim to simplify randomization through standardized workflows.

\section{Related work}
Our work is inspired by the observations around evaluation difficulties in deep RL in various prior lines of research. Henderson et al. observed how subtle implementation issues and random initialization drastically affect performance \cite{Henderson2017} across implementations. Mania et al. subsequently demonstrated that an augmented random search outperformed \cite{Mania2018} several policy optimization algorithms on supposedly difficult control tasks. Further recent work on policy gradient algorithms observed that the performance of popular algorithms may depend on implementation heuristics (e.g. learning rate annealing, reward normalization) which are not part of the core algorithm \cite{Ilyas2018}. For real-world RL, Dulac-Arnold et al. \cite{Dulac-Arnold2019} recently summarized  a set of nine core challenges needed to safely reach production.

In the wake of identifying evaluation challenges around Atari games, researchers have proposed specialized simulators to benchmark specific properties such as generalization capabilities (CoinRun \cite{Cobbe2018}) or agent safety (e.g. DeepMind safety gridworlds \cite{Leike2017}).  Bsuite \cite{Osband2019} is a novel benchmark for analyzing agent behaviour which varies random seeds to score agent performance but which does not distinguish different generalization modes or randomized tasks. 

To interface open source algorithm implementations, practitionerss have adopted OpenAI gym interfaces in novel simulators. For example, Siemens introduced a benchmark for industrial control tasks \cite{Hein2017a}. Others have built gym-bridges and new problem scenarios on to of existing simulators such as the ns3 networking simulator (ns3-gym \cite{Gawlowicz2018}). In systems-RL, Park is a benchmarking framework providing a common interface to a variety of problems in query processing, networking, or scheduling \cite{Mao2019a}. Park takes a first step towards shared task-design but does not explicitly include randomization and distinct blackbox and generalization modes.

\section{Conclusion}
We introduced Wield, a new tool towards systematic task construction and model evaluation for applied RL. Wield decouples application-specific protocols from task representation. We also introduced progressive randomization, an instructive evaluation protocol and classification scheme to analyze model capabilities under different randomization assumptions. Our assessment highlights the exciting recent progress in systems-RL, while demonstrating the substantial cost of delineating model capabilities.

\section*{Acknowledgments}
Michael Schaarschmidt is supported by a Google PhD Fellowship. We are also grateful for receiving research credits from Google Cloud. 

\bibliographystyle{sysml2019}

\begin{thebibliography}{0}
\providecommand{\natexlab}[1]{#1}
\providecommand{\url}[1]{\texttt{#1}}
\expandafter\ifx\csname urlstyle\endcsname\relax
  \providecommand{\doi}[1]{doi: #1}\else
  \providecommand{\doi}{doi: \begingroup \urlstyle{rm}\Url}\fi

\end{thebibliography}


\begin{thebibliography}{38}
\providecommand{\natexlab}[1]{#1}
\providecommand{\url}[1]{\texttt{#1}}
\expandafter\ifx\csname urlstyle\endcsname\relax
  \providecommand{\doi}[1]{doi: #1}\else
  \providecommand{\doi}{doi: \begingroup \urlstyle{rm}\Url}\fi

\bibitem[Addanki et~al.(2019)Addanki, Venkatakrishnan, Gupta, Mao, and
  Alizadeh]{Addanki2019}
Addanki, R., Venkatakrishnan, S.~B., Gupta, S., Mao, H., and Alizadeh, M.
\newblock Placeto: Learning generalizable device placement algorithms for
  distributed machine learning.
\newblock \emph{CoRR}, abs/1906.08879, 2019.
\newblock URL \url{http://arxiv.org/abs/1906.08879}.

\bibitem[Brockman et~al.(2016)Brockman, Cheung, Pettersson, Schneider,
  Schulman, Tang, and Zaremba]{openaigym}
Brockman, G., Cheung, V., Pettersson, L., Schneider, J., Schulman, J., Tang,
  J., and Zaremba, W.
\newblock Openai gym.
\newblock \emph{CoRR}, abs/1606.01540, 2016.
\newblock URL \url{http://arxiv.org/abs/1606.01540}.

\bibitem[Cobbe et~al.(2018)Cobbe, Klimov, Hesse, Kim, and Schulman]{Cobbe2018}
Cobbe, K., Klimov, O., Hesse, C., Kim, T., and Schulman, J.
\newblock Quantifying generalization in reinforcement learning.
\newblock \emph{arXiv preprint arXiv:1812.02341}, 2018.

\bibitem[Coleman et~al.(2018)Coleman, Kang, Narayanan, Nardi, Zhao, Zhang,
  Bailis, Olukotun, R{\'{e}}, and Zaharia]{Coleman2018}
Coleman, C., Kang, D., Narayanan, D., Nardi, L., Zhao, T., Zhang, J., Bailis,
  P., Olukotun, K., R{\'{e}}, C., and Zaharia, M.
\newblock Analysis of dawnbench, a time-to-accuracy machine learning
  performance benchmark.
\newblock \emph{CoRR}, abs/1806.01427, 2018.
\newblock URL \url{http://arxiv.org/abs/1806.01427}.

\bibitem[Dhariwal et~al.(2017)Dhariwal, Hesse, Klimov, Nichol, Plappert,
  Radford, Schulman, Sidor, Wu, and Zhokhov]{baselines}
Dhariwal, P., Hesse, C., Klimov, O., Nichol, A., Plappert, M., Radford, A.,
  Schulman, J., Sidor, S., Wu, Y., and Zhokhov, P.
\newblock Openai baselines.
\newblock \url{https://github.com/openai/baselines}, 2017.

\bibitem[Dulac-Arnold et~al.(2015)Dulac-Arnold, Evans, van Hasselt, Sunehag,
  Lillicrap, Hunt, Mann, Weber, Degris, and Coppin]{Dulac-Arnold2015}
Dulac-Arnold, G., Evans, R., van Hasselt, H., Sunehag, P., Lillicrap, T., Hunt,
  J., Mann, T., Weber, T., Degris, T., and Coppin, B.
\newblock Deep reinforcement learning in large discrete action spaces.
\newblock 2015.

\bibitem[Dulac{-}Arnold et~al.(2019)Dulac{-}Arnold, Mankowitz, and
  Hester]{Dulac-Arnold2019}
Dulac{-}Arnold, G., Mankowitz, D.~J., and Hester, T.
\newblock Challenges of real-world reinforcement learning.
\newblock \emph{CoRR}, abs/1904.12901, 2019.
\newblock URL \url{http://arxiv.org/abs/1904.12901}.

\bibitem[Gawlowicz \& Zubow(2018)Gawlowicz and Zubow]{Gawlowicz2018}
Gawlowicz, P. and Zubow, A.
\newblock ns3-gym: Extending openai gym for networking research.
\newblock \emph{CoRR}, abs/1810.03943, 2018.
\newblock URL \url{http://arxiv.org/abs/1810.03943}.

\bibitem[Guu et~al.(2017)Guu, Pasupat, Liu, and Liang]{Guu2017}
Guu, K., Pasupat, P., Liu, E.~Z., and Liang, P.
\newblock From language to programs: Bridging reinforcement learning and
  maximum marginal likelihood.
\newblock In Barzilay, R. and Kan, M. (eds.), \emph{Proceedings of the 55th
  Annual Meeting of the Association for Computational Linguistics, {ACL} 2017,
  Vancouver, Canada, July 30 - August 4, Volume 1: Long Papers}, pp.\
  1051--1062. Association for Computational Linguistics, 2017.
\newblock \doi{10.18653/v1/P17-1097}.

\bibitem[Hein et~al.(2017)Hein, Depeweg, Tokic, Udluft, Hentschel, Runkler, and
  Sterzing]{Hein2017a}
Hein, D., Depeweg, S., Tokic, M., Udluft, S., Hentschel, A., Runkler, T.~A.,
  and Sterzing, V.
\newblock A benchmark environment motivated by industrial control problems.
\newblock \emph{arXiv preprint arXiv:1709.09480}, 2017.

\bibitem[Henderson et~al.(2017)Henderson, Islam, Bachman, Pineau, Precup, and
  Meger]{Henderson2017}
Henderson, P., Islam, R., Bachman, P., Pineau, J., Precup, D., and Meger, D.
\newblock Deep reinforcement learning that matters.
\newblock \emph{CoRR}, abs/1709.06560, 2017.
\newblock URL \url{http://arxiv.org/abs/1709.06560}.

\bibitem[Ilyas et~al.(2018)Ilyas, Engstrom, Santurkar, Tsipras, Janoos,
  Rudolph, and Madry]{Ilyas2018}
Ilyas, A., Engstrom, L., Santurkar, S., Tsipras, D., Janoos, F., Rudolph, L.,
  and Madry, A.
\newblock Are deep policy gradient algorithms truly policy gradient algorithms?
\newblock \emph{CoRR}, abs/1811.02553, 2018.
\newblock URL \url{http://arxiv.org/abs/1811.02553}.

\bibitem[Jay et~al.(2019)Jay, Rotman, Godfrey, Schapira, and Tamar]{Jay2019}
Jay, N., Rotman, N.~H., Godfrey, B., Schapira, M., and Tamar, A.
\newblock A deep reinforcement learning perspective on internet congestion
  control.
\newblock In Chaudhuri, K. and Salakhutdinov, R. (eds.), \emph{Proceedings of
  the 36th International Conference on Machine Learning, {ICML} 2019, 9-15 June
  2019, Long Beach, California, {USA}}, volume~97 of \emph{Proceedings of
  Machine Learning Research}, pp.\  3050--3059. {PMLR}, 2019.
\newblock URL \url{http://proceedings.mlr.press/v97/jay19a.html}.

\bibitem[Johnson et~al.(2016)Johnson, Hofmann, Hutton, and
  Bignell]{Johnson2016}
Johnson, M., Hofmann, K., Hutton, T., and Bignell, D.
\newblock The malmo platform for artificial intelligence experimentation.
\newblock In \emph{IJCAI}, pp.\  4246--4247, 2016.

\bibitem[Juliani et~al.(2018)Juliani, Berges, Vckay, Gao, Henry, Mattar, and
  Lange]{Juliani2018a}
Juliani, A., Berges, V.-P., Vckay, E., Gao, Y., Henry, H., Mattar, M., and
  Lange, D.
\newblock Unity: A general platform for intelligent agents.
\newblock \emph{arXiv preprint arXiv:1809.02627}, 2018.

\bibitem[Kansky et~al.(2017)Kansky, Silver, M{\'e}ly, Eldawy,
  L{\'a}zaro-Gredilla, Lou, Dorfman, Sidor, Phoenix, and George]{Kansky2017}
Kansky, K., Silver, T., M{\'e}ly, D.~A., Eldawy, M., L{\'a}zaro-Gredilla, M.,
  Lou, X., Dorfman, N., Sidor, S., Phoenix, S., and George, D.
\newblock Schema networks: Zero-shot transfer with a generative causal model of
  intuitive physics.
\newblock 2017.

\bibitem[Leike et~al.(2017)Leike, Martic, Krakovna, Ortega, Everitt, Lefrancq,
  Orseau, and Legg]{Leike2017}
Leike, J., Martic, M., Krakovna, V., Ortega, P.~A., Everitt, T., Lefrancq, A.,
  Orseau, L., and Legg, S.
\newblock {AI} safety gridworlds.
\newblock \emph{CoRR}, abs/1711.09883, 2017.
\newblock URL \url{http://arxiv.org/abs/1711.09883}.

\bibitem[Leis et~al.(2015)Leis, Gubichev, Mirchev, Boncz, Kemper, and
  Neumann]{Leis2015}
Leis, V., Gubichev, A., Mirchev, A., Boncz, P., Kemper, A., and Neumann, T.
\newblock How good are query optimizers, really?
\newblock \emph{Proceedings of the VLDB Endowment}, 9\penalty0 (3):\penalty0
  204--215, 2015.

\bibitem[Liang et~al.(2018)Liang, Liaw, Nishihara, Moritz, Fox, Goldberg,
  Gonzalez, Jordan, and Stoica]{Liang2018}
Liang, E., Liaw, R., Nishihara, R., Moritz, P., Fox, R., Goldberg, K.,
  Gonzalez, J., Jordan, M., and Stoica, I.
\newblock Rllib: Abstractions for distributed reinforcement learning.
\newblock In \emph{International Conference on Machine Learning}, pp.\
  3059--3068, 2018.

\bibitem[Liang et~al.(2019)Liang, Zhu, Jin, and Stoica]{Liang2019}
Liang, E., Zhu, H., Jin, X., and Stoica, I.
\newblock Neural packet classification.
\newblock \emph{CoRR}, abs/1902.10319, 2019.
\newblock URL \url{http://arxiv.org/abs/1902.10319}.

\bibitem[Mania et~al.(2018)Mania, Guy, and Recht]{Mania2018}
Mania, H., Guy, A., and Recht, B.
\newblock Simple random search provides a competitive approach to reinforcement
  learning.
\newblock \emph{CoRR}, abs/1803.07055, 2018.
\newblock URL \url{http://arxiv.org/abs/1803.07055}.

\bibitem[Mao et~al.(2018)Mao, Schwarzkopf, Venkatakrishnan, Meng, and
  Alizadeh]{Mao2018b}
Mao, H., Schwarzkopf, M., Venkatakrishnan, S.~B., Meng, Z., and Alizadeh, M.
\newblock Learning scheduling algorithms for data processing clusters.
\newblock \emph{arXiv preprint arXiv:1810.01963}, 2018.

\bibitem[Mao et~al.(2019{\natexlab{a}})Mao, Negi, Narayan, Wang, Yang, Wang,
  Marcus, Addanki, Khani, He, et~al.]{Mao2019a}
Mao, H., Negi, P., Narayan, A., Wang, H., Yang, J., Wang, H., Marcus, R.,
  Addanki, R., Khani, M., He, S., et~al.
\newblock Park: An open platform for learning augmented computer systems.
\newblock \emph{Reinforcement Learning for Real Life Workshop, ICML},
  2019{\natexlab{a}}.
\newblock URL \url{https://github.com/park-project/park}.

\bibitem[Mao et~al.(2019{\natexlab{b}})Mao, Schwarzkopf, Venkatakrishnan, Meng,
  and Alizadeh]{Mao2019}
Mao, H., Schwarzkopf, M., Venkatakrishnan, S.~B., Meng, Z., and Alizadeh, M.
\newblock Learning scheduling algorithms for data processing clusters.
\newblock In Wu, J. and Hall, W. (eds.), \emph{Proceedings of the {ACM} Special
  Interest Group on Data Communication, {SIGCOMM} 2019, Beijing, China, August
  19-23, 2019}, pp.\  270--288. {ACM}, 2019{\natexlab{b}}.
\newblock \doi{10.1145/3341302.3342080}.

\bibitem[Marcus \& Papaemmanouil(2018)Marcus and Papaemmanouil]{Marcus2018}
Marcus, R. and Papaemmanouil, O.
\newblock Deep reinforcement learning for join order enumeration.
\newblock In Bordawekar, R. and Shmueli, O. (eds.), \emph{Proceedings of the
  First International Workshop on Exploiting Artificial Intelligence Techniques
  for Data Management, aiDM@SIGMOD 2018, Houston, TX, USA, June 10, 2018}, pp.\
   3:1--3:4. {ACM}, 2018.
\newblock \doi{10.1145/3211954.3211957}.

\bibitem[{Marcus} et~al.(2019){Marcus}, {Negi}, {Mao}, {Zhang}, {Alizadeh},
  {Kraska}, {Papaemmanouil}, and {Tatbul}]{2019arXiv190403711M}
{Marcus}, R., {Negi}, P., {Mao}, H., {Zhang}, C., {Alizadeh}, M., {Kraska}, T.,
  {Papaemmanouil}, O., and {Tatbul}, N.
\newblock {Neo: A Learned Query Optimizer}.
\newblock \emph{arXiv e-prints}, art. arXiv:1904.03711, Apr 2019.

\bibitem[Mirhoseini et~al.(2017)Mirhoseini, Pham, Le, Steiner, Larsen, Zhou,
  Kumar, Norouzi, Bengio, and Dean]{mirhoseini2017device}
Mirhoseini, A., Pham, H., Le, Q.~V., Steiner, B., Larsen, R., Zhou, Y., Kumar,
  N., Norouzi, M., Bengio, S., and Dean, J.
\newblock Device placement optimization with reinforcement learning.
\newblock \emph{arXiv preprint arXiv:1706.04972}, 2017.

\bibitem[Mirhoseini et~al.(2018)Mirhoseini, Goldie, Pham, Steiner, Le, and
  Dean]{hierarchical2018}
Mirhoseini, A., Goldie, A., Pham, H., Steiner, B., Le, Q.~V., and Dean, J.
\newblock Hierarchical planning for device placement.
\newblock 2018.
\newblock URL \url{https://openreview.net/pdf?id=Hkc-TeZ0W}.

\bibitem[OpenAI(2018)]{openaidota}
OpenAI.
\newblock {OpenAI Five DOTA}.
\newblock Website, June 2018.
\newblock URL \url{https://blog.openai.com/openai-five/}.

\bibitem[Ortiz et~al.(2018)Ortiz, Balazinska, Gehrke, and Keerthi]{Ortiz2018}
Ortiz, J., Balazinska, M., Gehrke, J., and Keerthi, S.~S.
\newblock Learning state representations for query optimization with deep
  reinforcement learning.
\newblock In Schelter, S., Seufert, S., and Kumar, A. (eds.), \emph{Proceedings
  of the Second Workshop on Data Management for End-To-End Machine Learning,
  DEEM@SIGMOD 2018, Houston, TX, USA, June 15, 2018}, pp.\  4:1--4:4. {ACM},
  2018.
\newblock \doi{10.1145/3209889.3209890}.

\bibitem[Osband et~al.()Osband, Doron, Hessel, Aslanides, Sezener, Saraiva,
  McKinney, Lattimore, Szepezvari, Singh, Roy, Sutton, Silver, and
  Hasselt]{Osband2019}
Osband, I., Doron, Y., Hessel, M., Aslanides, J., Sezener, E., Saraiva, A.,
  McKinney, K., Lattimore, T., Szepezvari, C., Singh, S., Roy, B.~V., Sutton,
  R., Silver, D., and Hasselt, H.~V.
\newblock Behaviour suite for reinforcement learning.

\bibitem[Paliwal et~al.(2019)Paliwal, Gimeno, Nair, Li, Lubin, Kohli, and
  Vinyals]{Paliwal2019}
Paliwal, A., Gimeno, F., Nair, V., Li, Y., Lubin, M., Kohli, P., and Vinyals,
  O.
\newblock {REGAL:} transfer learning for fast optimization of computation
  graphs.
\newblock \emph{CoRR}, abs/1905.02494, 2019.
\newblock URL \url{http://arxiv.org/abs/1905.02494}.

\bibitem[Schaarschmidt et~al.(2019)Schaarschmidt, Mika, Fricke, and
  Yoneki]{Schaarschmidt2019}
Schaarschmidt, M., Mika, S., Fricke, K., and Yoneki, E.
\newblock {RLgraph: Modular Computation Graphs for Deep Reinforcement
  Learning}.
\newblock In \emph{{Proceedings of the 2nd Conference on Systems and Machine
  Learning (SysML)}}, April 2019.

\bibitem[Schulman et~al.()Schulman, Abbeel, and Chen]{Schulman2017}
Schulman, J., Abbeel, P., and Chen, X.
\newblock Equivalence between policy gradients and soft q-learning.

\bibitem[Silver et~al.(2018)Silver, Hubert, Schrittwieser, Antonoglou, Lai,
  Guez, Lanctot, Sifre, Kumaran, Graepel, Lillicrap, Simonyan, and
  Hassabis]{Silver1140}
Silver, D., Hubert, T., Schrittwieser, J., Antonoglou, I., Lai, M., Guez, A.,
  Lanctot, M., Sifre, L., Kumaran, D., Graepel, T., Lillicrap, T., Simonyan,
  K., and Hassabis, D.
\newblock A general reinforcement learning algorithm that masters chess, shogi,
  and go through self-play.
\newblock \emph{Science}, 362\penalty0 (6419):\penalty0 1140--1144, 2018.
\newblock ISSN 0036-8075.
\newblock \doi{10.1126/science.aar6404}.
\newblock URL \url{https://science.sciencemag.org/content/362/6419/1140}.

\bibitem[Sutton et~al.(1999)Sutton, Precup, and Singh]{Sutton1999}
Sutton, R.~S., Precup, D., and Singh, S.~P.
\newblock Between mdps and semi-mdps: {A} framework for temporal abstraction in
  reinforcement learning.
\newblock \emph{Artif. Intell.}, 112\penalty0 (1-2):\penalty0 181--211, 1999.
\newblock \doi{10.1016/S0004-3702(99)00052-1}.

\bibitem[Tesauro et~al.(2006)Tesauro, Jong, Das, and
  Bennani]{TesauroJongDasEtAl2006}
Tesauro, G., Jong, N.~K., Das, R., and Bennani, M.~N.
\newblock A hybrid reinforcement learning approach to autonomic resource
  allocation.
\newblock In \emph{Proceedings of the 2006 IEEE International Conference on
  Autonomic Computing}, ICAC '06, pp.\  65--73, Washington, DC, USA, 2006. IEEE
  Computer Society.
\newblock ISBN 1-4244-0175-5.
\newblock \doi{10.1109/ICAC.2006.1662383}.
\newblock URL \url{http://dx.doi.org/10.1109/ICAC.2006.1662383}.

\bibitem[Valadarsky et~al.(2017)Valadarsky, Schapira, Shahaf, and
  Tamar]{Valadarsky2017}
Valadarsky, A., Schapira, M., Shahaf, D., and Tamar, A.
\newblock A machine learning approach to routing.
\newblock \emph{arXiv preprint arXiv:1708.03074}, 2017.

\end{thebibliography}
\newpage

\appendix
\section{Experiment hyperparameters}
We list all hyperparameters used in Wield's hierarchical placer. Table \ref{grouper-hparams} lists grouper parameters.

\begin{table}[h]
  \caption{Training parameters used for Wield's grouper agent.}
  \label{grouper-hparams}
  \centering
  \begin{tabular}{ll}
    \cmidrule{1-2}
   Parameter     & Value \\
    \midrule
clip ratio & $0.25$ \\
discount & $1.0$ \\
GAE $\lambda$ & $1.0$ \\
batch size & num ops in graph \\
update iterations per batch & $10$ \\
num groups & $20$ \\
policy layer size & $32$ \\
num hidden layers & $2$ \\
layer activation & tanh \\
value function & same as policy \\
optimizer & Adam \\
learning rate & $[0.01, 0.00001]$ \\
linear decay steps & $600$ \\
  \end{tabular}
\end{table}

Table \ref{placer-hparams} lists placer parameters. The embedding implementation is a faithful implementation of Addanki et al.'s description \cite{Addanki2019}. The main difference is that their work relies on manual grouping. When we tested initially random groupings produced by the grouper, neighborhood embeddings diverged due to large neighborhood sets. We hence used \textit{tanh} activations instead of rectified linear units, and configured the grouper to a more aggressive learning rate schedule. 

\begin{table}[h]
  \caption{Training parameters used for Wield's placer agent.}
  \label{placer-hparams}
  \centering
  \begin{tabular}{ll}
    \cmidrule{1-2}
   Parameter     & Value \\
    \midrule
clip ratio & $0.2$ \\
discount & $1.0$ \\
GAE $\lambda$ & $1.0$ \\
batch size & $10$ \\
update iterations per batch & $1$ \\
layer size (all  layers) & num groups \\
num in neighbors & $5$ \\
num out neighbors & $5$ \\
neighborhood aggregation rounds & $10$ \\
layer activation & tanh \\
value function & same as policy \\
optimizer & Adam \\
learning rate & $[0.0003, 0.0001]$ \\
linear decay steps & $1000$ \\
groups placed per graph evaluation & $10$ \\
  \end{tabular}
\end{table}

~ \\

The NMT architecture was taken from Google's NMT implementation\footnote{https://github.com/tensorflow/nmt}. We used the 'normed\_bahdanau' attention layer with the 'gnmt\_v2' architecture.

\end{document}